\documentclass{article}

\usepackage[utf8]{inputenc} 
\usepackage[T1]{fontenc}    
\usepackage{hyperref}       
\usepackage{url}            
\usepackage{booktabs}       
\usepackage{amsfonts}       
\usepackage{nicefrac}       
\usepackage{microtype}      
\usepackage{xcolor}         
\usepackage{graphicx}
\usepackage{subcaption}
\usepackage{caption}
\usepackage{hyphenat}
\usepackage{setspace}
\usepackage{amsmath}
\usepackage{amssymb}
\usepackage{mathtools}
\usepackage{amsthm}
\usepackage{longtable}
\usepackage{multirow}
\usepackage{float}
\usepackage{tabularx}
\usepackage{enumitem}
\usepackage{threeparttable}
\usepackage{algorithm}
\usepackage{algorithmic}
\usepackage{titletoc}

\usepackage[preprint]{my_style}

\usepackage{xspace}

\graphicspath{{figs/}{figures/}}

\newcommand{\ours}{IndustryForge-27B\xspace}
\newcommand{\qwenM}{Qwen3.5-27B\xspace}
\newcommand{\gptM}{gpt-5.4\xspace}

\usepackage{tcolorbox}
\tcbuselibrary{skins, breakable, listings}

\definecolor{brandblue}{RGB}{57,95,207}
\definecolor{linkblue}{HTML}{0064E0}
\definecolor{textgray}{HTML}{1C2B33}
\definecolor{boxbg}{HTML}{F1F4F7}

\hypersetup{
  colorlinks=true,
  linkcolor=brandblue,
  citecolor=brandblue,
  urlcolor=linkblue
}

\newcommand{\paperTitle}{IndustryForge-27B: A Domain-Enhanced Multimodal Foundation Model for Industrial CAD}

\newcommand{\paperAuthors}{%
  {\sffamily\bfseries
    Nianchen Deng$^{1}$,\
    Jiaxin Ai$^{2,1,3}$,\
    Tao Hu$^{4,1}$,\
    Shu Zou$^{1,5}$,\
    Yurui Dong$^{1,6}$,\
    Siqi Li$^{1,7}$,\
    Xinyu Cai$^{1}$,\
    Xuemeng Yang$^{1}$,\
    Licheng Wen$^{1,3}$,\
    Hongbin Zhou$^{1}$,\
    Hairong Zhang$^{6,1}$,\
    Pinlong Cai$^{1,*}$,\
    Botian Shi$^{1,3}$%
  }%
}

\newcommand{\paperAffiliations}{%
  {\normalsize
    $^{1}$ Shanghai Artificial Intelligence Laboratory\quad
    $^{2}$ Wuhan University\quad
    $^{3}$ Shanghai Innovation Institute\\
    $^{4}$ University of Science and Technology of China\quad
    $^{5}$ The Australian National University\quad
    $^{6}$ Fudan University\quad
    $^{7}$ Zhejiang University%
  }%
}

\newcommand{\paperNotes}{%
  {\small $^{*}$ Corresponding author.}%
}

\newcommand{\hfLink}{\url{https://huggingface.co/Nianchen/IndustryForge-27B}}
\newcommand{\publishDate}{\today}

\newcommand{\renderFrontBox}{%
    \tcbset{
    enhanced, frame hidden,
    colback=boxbg,
    left=0.5cm, right=0.5cm, top=0.5cm, bottom=0.5cm,
    arc=16pt,
    before skip=0pt,
    grow to left by=1.5pt, grow to right by=1.5pt,
    overlay={
    \node[anchor=north east, at=(frame.north east), xshift=-2.3cm, yshift=-0.5cm] 
        {\includegraphics[height=1cm]{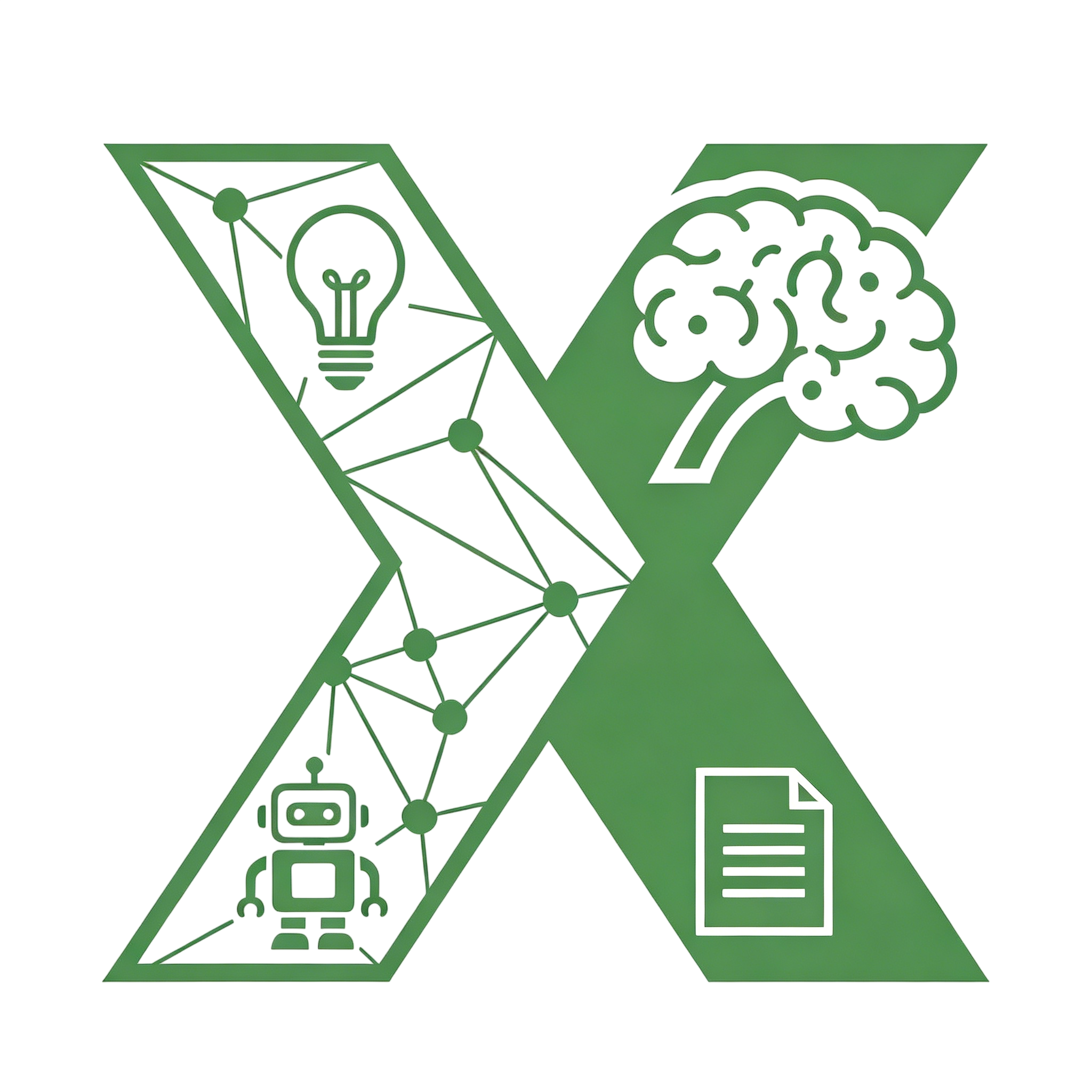}};
    \node[anchor=north east, at=(frame.north east), xshift=-0.5cm, yshift=-0.5cm] 
        {\includegraphics[height=1cm]{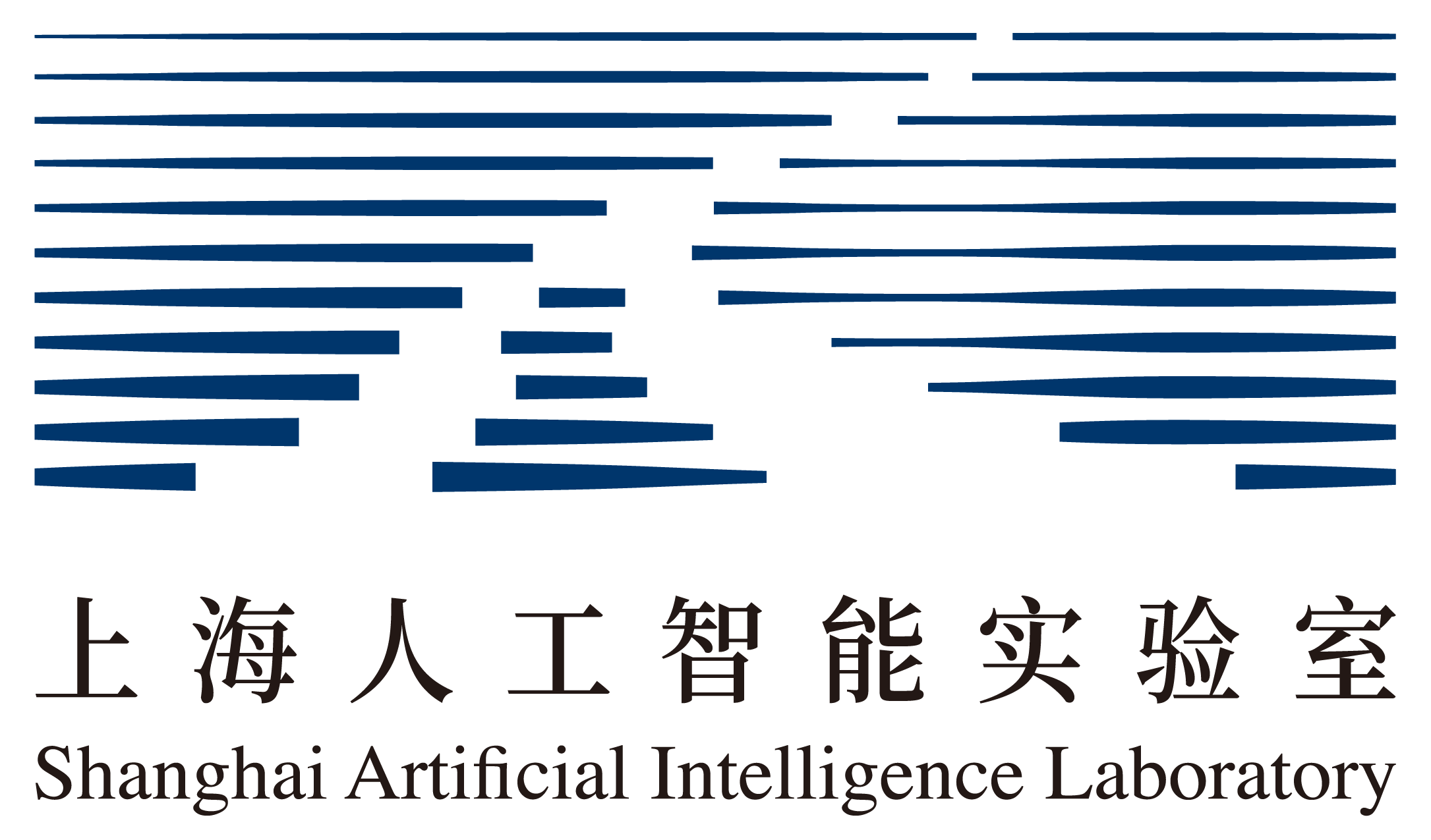}};
    }
  }%
  \begin{tcolorbox}
    \setlength{\parindent}{0cm}
    \setlength{\parskip}{0.5cm}
    {
      \setlength{\parskip}{0cm}
      \raggedright
      \nohyphens
      {
        \vskip 1cm 
        \setstretch{1.4}
        {\huge\sffamily\bfseries\textcolor{black}{\paperTitle}}\par
      }
      \vskip 0.25cm
      \paperAuthors\par
      \vskip 0.35cm
      \paperAffiliations\par
      \vskip 0.08cm
      \paperNotes\par
    }
    \vskip 0.2cm
    {\color{textgray}%
Automating industrial CAD design and manufacturing places distinctive demands on
multimodal foundation models: the model must \emph{see} engineering drawings and
3D geometry screenshots, \emph{write} correct parametric-modelling scripts and
Windows COM API code, and cover the full range from single parts to assemblies.
General-purpose multimodal models fall short on these tasks, while single-task
fine-tuning is too narrow to support the diverse calls that upper-layer agents
issue. We build \textbf{IndustryForge-27B} on top of Qwen3.5-VL-27B by curating
and integrating six industrial-CAD sub-corpora totalling $\sim$52k multimodal
samples---CAD Visual QA (CAD-VQA), parametric CAD code (text2cadquery),
assembly-level CAD code (text2cadquery-assembly), and three COM sub-corpora for
Inventor / SolidWorks (com\_2d / com\_3d / com\_assembly)---and training with a
unified multi-task SFT recipe. Across four CAD-domain benchmarks \ours
lifts the base model by $+33.65$~pp on average and \emph{outperforms} the strong
closed-source model \gptM on all four; across eleven general-capability
benchmarks it retains, and slightly improves upon, the base model ($+1.56$~pp
mean, no catastrophic forgetting). \ours \emph{will} serve as the
common substrate for downstream industrial-agent projects, providing a
unified starting point for a full-stack industrial agent that spans from CAD
design to industrial-software operation, from parts to assemblies, and from
single-shot generation to closed-loop self-improvement.

\par}
    \vskip 0.4cm
    {
      \setlength{\parskip}{0cm}
      {\small {\sffamily\bfseries \raisebox{-0.2em}{\includegraphics[width=0.025\linewidth]{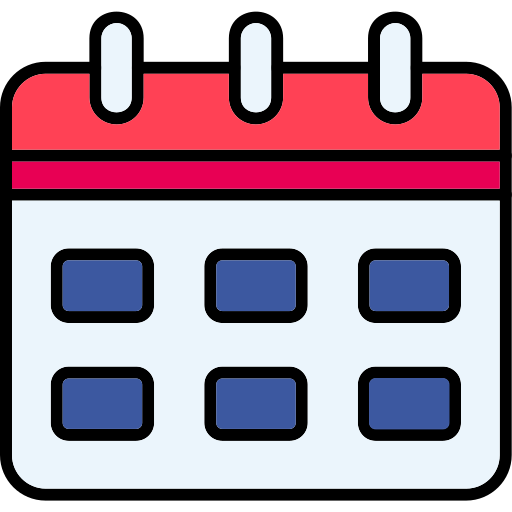}}~~Date:} \publishDate}\par%
      \vskip 0.08cm%
      {\small {\sffamily\bfseries \raisebox{-0.2em}{\includegraphics[width=0.025\linewidth]{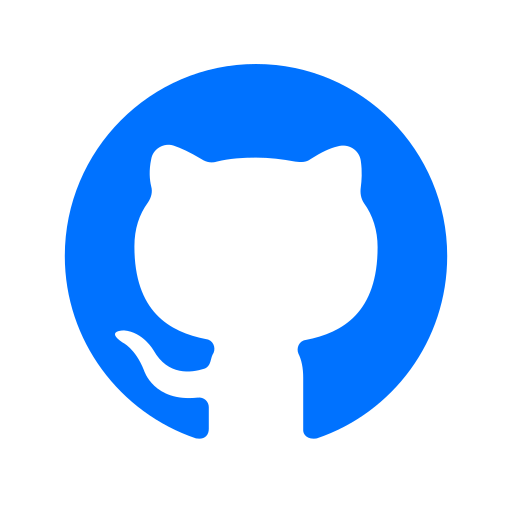}}~~Hugging Face Model:} \hfLink}
    }
  \end{tcolorbox}
  \tcbset{reset}
}

\begin{document}

\newgeometry{top=1in, bottom=0.75in, textwidth=6.3in, textheight=9in}
\renderFrontBox

\begin{figure}[H]
  \centering
  \includegraphics[width=\columnwidth]{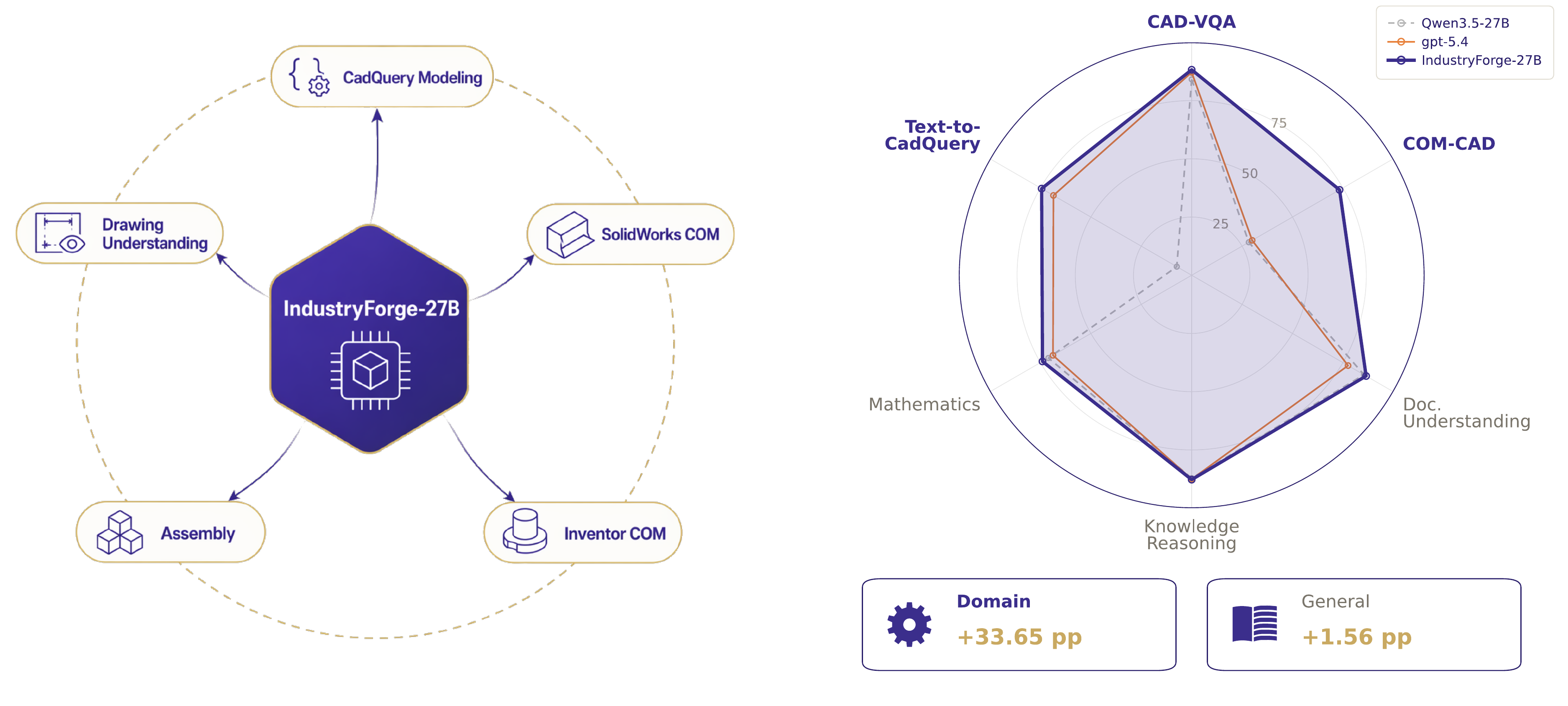}
  \caption{\textbf{\ours at a glance.} \emph{Left --- Capability Radiation.}
  Five industrial-CAD skills (Drawing Understanding, CadQuery Modeling,
  SolidWorks COM, Inventor COM, Assembly) radiate from the \ours badge
  and share a common ``full-stack CAD loop'' dashed orbit that visualises the
  end-to-end \emph{design $\to$ modeling $\to$ assembly $\to$
  software-operation} pipeline the substrate supports. \emph{Right --- Six-axis
  radar.} Three domain axes (upper half: CAD-VQA, Text-to-CadQuery, COM-CAD)
  and three aggregated general-capability axes (lower half: Doc.\
  Understanding, Knowledge Reasoning, Mathematics) compare \ours
  (blue-purple, filled) against \qwenM (grey, dashed) and \gptM (orange).
  Two floating cards report the headline: \ours wins 4/4 CAD-domain
  benchmarks against \gptM ($+33.65$~pp mean) while retaining general
  capabilities on 11 public benchmarks ($+1.56$~pp mean, no catastrophic
  forgetting).}
  \label{fig:teaser}
\end{figure}


\section{Introduction}
\label{sec:intro}

\subsection{Why industrial design and manufacturing need a specialised foundation model}

Computer-Aided Design (CAD) sits at the starting point of virtually every
modern manufacturing pipeline: from a single screw to an entire aircraft,
engineers carve parts, assemblies, and engineering drawings out of parametric
modelling software one operation at a time. In the past two years, the rise of
\emph{programmatic CAD}~\citep{cadquery,openscad} together with the rapid
improvement of large language models on code generation has made it feasible
to obtain executable CAD scripts from a natural-language description or a
single rendered image~\citep{deepcad,text2cad}. Nevertheless, once we push
beyond ``a program that runs'' towards ``a manufacturable, assemblable
engineering artefact'', general-purpose multimodal foundation models
degrade sharply along three axes:

\begin{itemize}
    \item \textbf{``Reads the drawing, writes the wrong code''.} A general
    multimodal model can caption an isometric screenshot in prose that sounds
    plausible, yet when the same content has to be translated into CadQuery
    or OpenSCAD scripts, dimensions, constraints, and coordinate frames are
    frequently mis-assigned---the shape is qualitatively right but the numbers
    are entirely wrong.

    \item \textbf{Poor coverage of industrial APIs.} Windows-COM interfaces
    exposed by Inventor, SolidWorks, and AutoCAD occupy a vanishing share of
    public code corpora. General models have essentially never seen how to
    \emph{extrude a feature via COM}, let alone how to assemble parts through
    \texttt{InsertMoveCopyBody2} and mate constraints.

    \item \textbf{Absence of the assembly scale.} Public CAD data are
    overwhelmingly single-part; assembly-level constraints, coaxial-alignment,
    and interference checking are chronically under-represented, so
    general-purpose models score close to zero on multi-part systems
    (\S\ref{sec:domain-results}).
\end{itemize}

None of these shortcomings can be patched by any single downstream agent
alone; they must be addressed at the \emph{foundation-model layer}. That is
the starting point of this work.

\subsection{Three limitations of existing routes}

\paragraph{(1) Single-task fine-tuning.}
Systems in the DeepCAD / SkexGen / Text2CAD lineage~\citep{deepcad,skexgen,text2cad}
typically over-fit one narrow task---for example ``text $\to$ single-part sketch
sequence''---and generalise poorly across the multiple call patterns that a
higher-level agent must dispatch.

\paragraph{(2) Prompting closed-source models directly.}
Using a strong closed-source model as the sole worker in a real engineering
pipeline is expensive, difficult to control, and hits the coverage ceiling
described above: closed-source models are dramatically weaker on Inventor /
SolidWorks COM code, and near-zero on assembly-level tasks
(\gptM scores $3.85\%$ on CadQuery Assembly in \S\ref{sec:domain-results}).

\paragraph{(3) No unified foundation.}
Capabilities within the CAD ecosystem---from part to assembly, from CadQuery
to COM---are currently scattered across specialised subsystems: IterCAD
targets closed-loop generation, AssemCAD targets assemblies, ComAct targets
COM operations, and SimLoop targets the CAD-CAE loop. These subsystems share
no common \emph{substrate}, forcing each team to independently solve the
prerequisite question ``can our foundation model read CAD drawings and write
CAD code at all?''---a large and duplicated up-front cost.

\subsection{Our stance: the foundation model as common substrate}
\label{sec:our-stance}

\ours does not attempt to solve every industrial-agent problem
end-to-end. Instead, we abstract four \emph{shared skills} out of the
industrial-CAD landscape and pre-install all four into a single 27B-parameter
multimodal model via multi-task SFT: (i)~\textbf{CAD visual understanding},
(ii)~\textbf{CadQuery parametric code generation},
(iii)~\textbf{COM API code generation} for Inventor / SolidWorks / AutoCAD, and
(iv)~\textbf{assembly-level CAD code generation}. Concretely:

\begin{quote}
\itshape \ours is a common substrate for industrial agents,
not an end-to-end CAD generation system.
\end{quote}

Figure~\ref{fig:teaser} summarises the resulting positioning. The
\emph{left} panel radiates the four shared skills as five capability
tags---Drawing Understanding, CadQuery Modeling, SolidWorks COM, Inventor
COM (COM is split by target application), and Assembly---out of the
\ours badge, all sharing a common ``full-stack CAD loop'' orbit that
visualises the end-to-end design $\to$ modeling $\to$ assembly $\to$
software-operation pipeline the substrate supports. The \emph{right} panel
reports headline scores on a six-axis radar (three domain axes plus three
general-capability aggregates) against \qwenM and \gptM: \ours
wins on every domain axis while preserving---and slightly improving---its
general capabilities.

\subsection{Contributions}

\begin{itemize}
    \item \textbf{An industrial multimodal SFT corpus.} We curate and integrate
    four task families over six sub-corpora, $\sim$52k samples in total
    (\S\ref{sec:data-construction}). The three COM sub-corpora share their
    upstream with ComAct~\citep{comact}, text2cadquery shares its upstream with
    IterCAD~\citep{itercad}, and text2cadquery-assembly with
    AssemCAD~\citep{assemcad}; the CAD-VQA sub-corpus is independently
    constructed by a collaborator on this project \emph{[TODO: source and pipeline
    for cad\_vqa to be filled in by the collaborator]}. All code samples are
    filtered through sandbox execution and, where applicable, Chamfer-Distance
    quality control.

    \item \textbf{A training recipe for unbalanced multi-task SFT.} On top of
    Qwen3.5-VL-27B we combine LoRA, DeepSpeed ZeRO-3, sequence-parallel~$=8$,
    and padding-free packing (\S\ref{sec:training-recipe}) with per-subset
    validation, producing a v4.1 configuration that trains stably on
    $8{\times}$A100-80\,GB.

    \item \textbf{A systematic evaluation suite for CAD code generation.} We
    put together a four-benchmark domain suite (CAD~VQA, CadQuery, COM~CAD,
    CadQuery~Assembly) and an eleven-benchmark general suite (aime26, arc,
    chartqa, gsm8k, gsm8k\_v, math\_vision, mmlu, mmmu, ocr\_bench, science\_qa,
    trivia\_qa), and report head-to-head comparisons against \qwenM and
    \gptM (\S\ref{sec:experiments}).

\end{itemize}

\section{Related Work}
\label{sec:related-work}

\subsection{Programmatic CAD}

The premise of \emph{programmatic CAD} is to describe a 3D model with executable
code rather than GUI clicks. CadQuery~\citep{cadquery}, OpenSCAD~\citep{openscad}
and Onshape FeatureScript are representative frameworks in this line: an
``extrude'' button is replaced by an equivalent Python or DSL statement, so
that the entire modelling process becomes version-controllable, parametric, and
automatable. This paradigm shift is especially favourable for large language
models---they no longer need to imitate pixel-level GUI interaction and can
instead work in their strongest medium, the code space. The CadQuery portion of
\ours rests directly on this code-driven CAD assumption.

\subsection{Text- and Image-to-CAD}

From early systems that emit sketch sequences
(DeepCAD~\citep{deepcad}, SkexGen~\citep{skexgen}), to more recent work whose
core task is ``natural language $\to$ CAD code or parameter sequence''
(Text2CAD~\citep{text2cad}, CAD-LLM~\citep{cadllm}), Text- and Image-to-CAD
has been an active research direction. In the past year, as code LLMs have
matured, the sub-line that generates CadQuery / OpenSCAD code directly has
attracted substantial attention. On top of this, a family of ``CAD agents''
aimed at real engineering pipelines has emerged:

\begin{itemize}
    \item \textbf{IterCAD}~\citep{itercad} lifts CAD code generation from a
    single-shot mapping into a ``generate $\to$ verify $\to$ refine''
    multi-round loop, combining compiler feedback, sandbox execution, and
    visual critique so that a 4B model can outperform closed-source giants
    such as GPT-5 on error rate and AUC-TR.

    \item \textbf{AssemCAD}~\citep{assemcad} targets Production-Ready Assembly
    Generation: it decomposes mechanical assemblies into ``What to Build /
    How to Build / How to Verify'' stages, letting the semantic reasoning of
    LLMs work in concert with a deterministic CAD kernel.

    \item \textbf{ComAct}~\citep{comact} introduces the
    \emph{COM-as-Action} paradigm: rather than have an agent stare at pixels
    and click buttons, ComAct operates SolidWorks / Inventor / AutoCAD's
    internal objects through COM code, paired with the ComCADBench benchmark
    and the ComForge parallel-execution platform for agent training.
\end{itemize}

These works each tackle a different slice of the CAD-agent problem, but they
share one prerequisite: \emph{the underlying model must first be able to read
CAD-relevant vision and to write at least one form of CAD code}.
\ours is designed as the ``qualified starting point'' for exactly
these downstream agents.

\subsection{Industrial software: the COM / API ecosystem}

Within the Windows ecosystem, the Component Object Model (COM) is a
system-level object model adopted by a large swath of traditional heavy
professional software---Autodesk Inventor, Dassault SolidWorks, Autodesk
AutoCAD, Microsoft Office, Adobe Suite, and many more. Compared to GUI click
chains or fragmented REST APIs, COM offers a much closer approximation of
the software's ``native language'': an agent directly manipulates sketches,
parts, and assemblies as internal objects, rather than emulating a mouse.
COM corpora, however, are extremely scarce on the public internet---most
publicly available tutorials cover only Office Automation, while CAD-specific
COM interfaces remain effectively locked away behind closed proprietary
documentation and internal scripts. Through three ComAct-aligned COM
sub-corpora---\texttt{com\_2d}, \texttt{com\_3d}, \texttt{com\_assembly},
totalling 30k samples---\ours substantially alleviates this
data scarcity.

\subsection{Multimodal foundation models}

The Qwen-VL series (Qwen-VL, Qwen2-VL, Qwen3-VL,
Qwen3.5-VL)~\citep{qwenvl,qwen25vl} and the InternVL
series~\citep{internvl} are representative open-source multimodal foundation
models of the past two years. Medicine, coding, and mathematics all offer
mature examples of domain SFT / CPT on such foundations. \ours
follows this ``general multimodal foundation + domain SFT'' recipe, extending
it from the text-code domain into the compound modality of
\emph{industrial-CAD vision + code}.

\paragraph{Positioning of this work.} \ours sits at the
\emph{foundation-model layer} between general multimodal foundation and
industrial-agent systems---deeper than DeepCAD / Text2CAD, because it is not
tied to any single downstream task, yet more specialised than closed-source
generalists, because it pre-injects CAD vision, CadQuery, COM, and assembly
capabilities.

\section{Methodology}
\label{sec:methodology}

\subsection{Overview}
\label{sec:overview}

\begin{figure}[t]
    \centering
    \includegraphics[width=0.98\columnwidth]{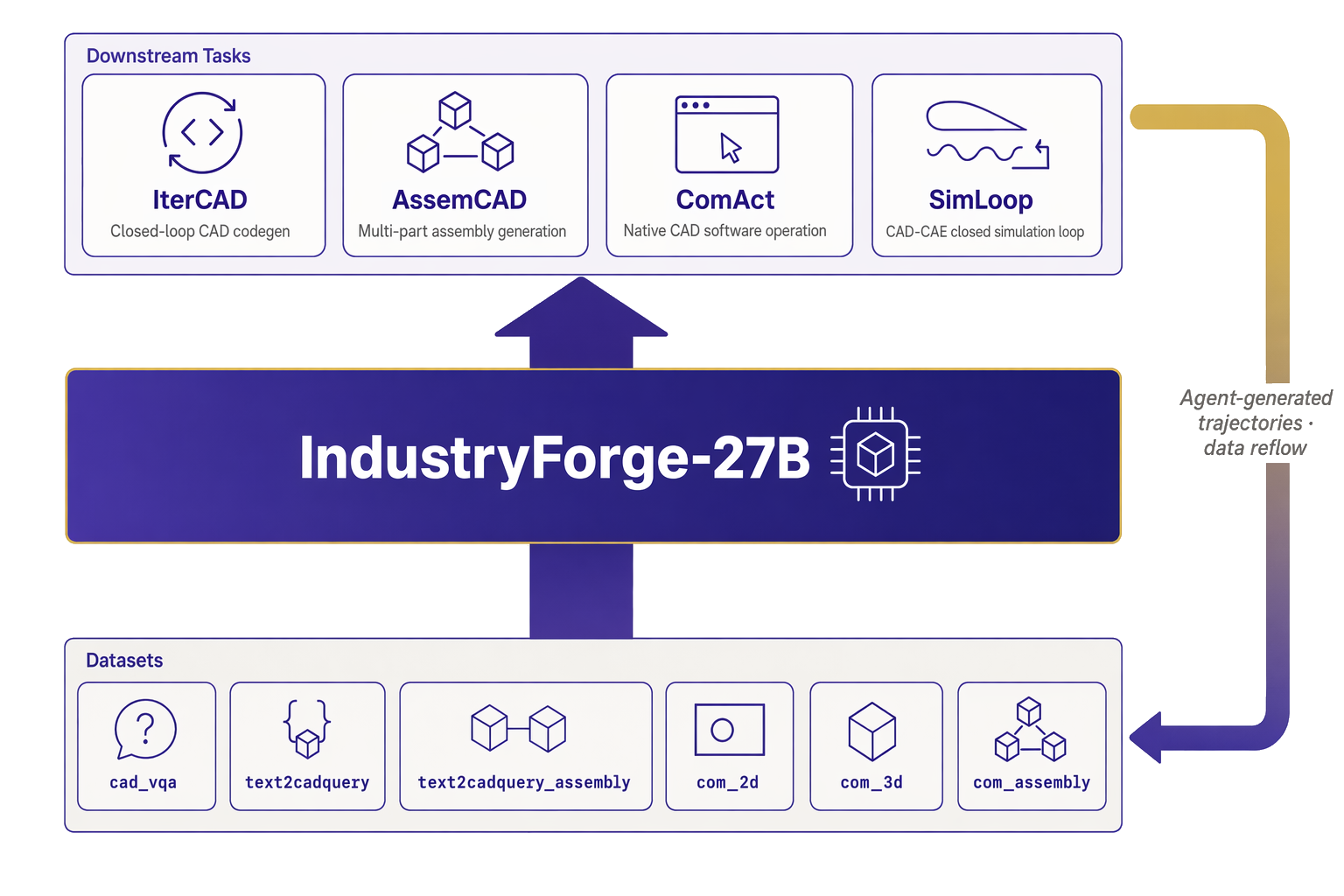}
    \caption{\textbf{Framework overview.} A vertical three-tier stack. \emph{Bottom:}
    six domain sub-corpora (CAD-VQA 4k, text2cadquery 17k, text2cadquery-assembly
    1k, com\_2d 20k, com\_3d 5k, com\_assembly 5k), colour-coded by their upstream
    project (IterCAD / AssemCAD / ComAct / independent). \emph{Middle:} the
    \ours foundation model, itself built on Qwen3.5-VL-27B and
    trained on $\sim$52k samples. \emph{Top:} four downstream agents that will
    consume the substrate---IterCAD (closed-loop generation), AssemCAD (assembly
    generation), ComAct (COM operation), and SimLoop (CAD-CAE loop). The main
    flow (wide arrow) is bottom-up. A thin \emph{return} arrow on the right
    marks a planned data-feedback path: high-quality traces produced by
    downstream agents in real tasks can be recycled into the substrate's next
    training round.}
    \label{fig:framework}
\end{figure}

The construction of \ours decomposes into three stages:
\emph{data} $\to$ \emph{SFT} $\to$ \emph{downstream agents}. Figure~\ref{fig:framework}
illustrates the resulting three-tier structure.

\subsection{Training data construction}
\label{sec:data-construction}

This section is the heart of the report. We organise the discussion around
four task families over six sub-corpora: CAD visual understanding, parametric
CAD code generation, assembly-level CAD code generation, and industrial-API
(COM) code generation.

\subsubsection{CAD visual understanding: CAD-VQA-4k}
\label{sec:cadvqa}

\paragraph{Objective.} Inject a ``CAD-reading'' prior into the foundation
model. Multi-view engineering drawings, isometric 3D renders, and structured
dimensional annotations occupy a negligible share of general-purpose
pre-training corpora, so the model must be explicitly taught to read the
geometry and drafting semantics carried by CAD-specific visual
representations.

\paragraph{Source and construction.} First, complex CAD drawings containing part drawings, assembly drawings and multi-view information are screened from raw mechanical drawings. In the question design phase, typical exam questions from university mechanical engineering courses including Mechanical Manufacturing Technology, Tolerances and Technical Measurements, and Fundamentals of Mechanical Design are systematically sorted out. Based on these materials, task categories are summarized, such as basic dimension reading, machining process planning, operating principles of mechanisms, assembly tolerance analysis, and machining error calculation. Afterwards, a multi-modal model is adopted to generate corresponding questions, standard answers and explanations. A review model is then used to verify the rationality of questions and the correctness of answers; unqualified samples are eliminated or revised. Finally, a structured CAD-VQA dataset is constructed, which includes images, questions, answers, explanations and question type labels.


\subsubsection{Parametric CAD code generation: text2cadquery-17k}
\label{sec:t2cq}

\paragraph{Objective.} Teach the model to translate a natural-language
description or a single-view render into executable \textbf{CadQuery} Python
code that renders to the target geometry under a deterministic kernel.

\paragraph{Source and construction.} This sub-corpus is co-sourced with the
$D_{S1}$ (drawing $\to$ code) stage of IterCAD's expert trajectories.
Raw annotations, drawn from normalised scripts in a programmatic-CAD library,
pass through the following pipeline to yield 17k high-quality samples:
(1)~sandbox execution verifies that every script runs and renders;
(2)~a Chamfer-Distance threshold filters out samples that ``run but drift
away from the intended geometry'', suppressing annotation noise;
(3)~each script is automatically paired with multi-view renders and a
natural-language description, forming a complete
$(\text{multimodal input} \to \text{code})$ QA pair; and (4)~the whole set
is deduplicated against every downstream benchmark at both UUID and
structural levels (\S\ref{sec:qc}). A more complete write-up of the underlying data curation
is available in the IterCAD paper (\S\,``Data Curation'').


\subsubsection{Assembly-level CAD code generation: text2cadquery-assembly-1k}
\label{sec:t2cq-asm}

\paragraph{Objective.} Push CAD code generation from ``single part'' to
``assembly''. A real mechanical system is made of multiple parts glued
together by engineering-level constraints, coaxial alignments, and
interference checks. A runnable piece of CAD code is far from sufficient to
guarantee an assemblable mechanical system.

\paragraph{Source and construction.} This sub-corpus is co-sourced with the
assembly data of AssemCAD. Construction follows AssemCAD's
What~/~How~/~Verify three-stage recipe: (1)~a large model first parses a
natural-language assembly requirement into a structured
\emph{Assembly Specification} that explicitly encodes the parts, interfaces,
and assembly relations; (2)~parametric part geometry and an assembly script
are generated from the Specification; and (3)~geometric interference checks,
assembly connectivity analysis, and engineering-rule validation gate the
output at the system level. Only assemblies that clear these checks enter
the SFT set. Because the acceptance bar is high, the final size lands at
$\sim$1k---small in count, but every sample carries an assembly-level
geometric-consistency guarantee. The full construction and verification pipeline is described
in the AssemCAD paper (Design Understanding and Assembly Construction).


\subsubsection{Industrial-software API code generation (the COM family)}
\label{sec:com}

\paragraph{Objective.} Equip the substrate with the ability to operate
mainstream industrial CAD software (SolidWorks / Inventor / AutoCAD), and to
do so \emph{not} through GUI click chains but through COM code that
directly invokes the software's internal objects---the
\emph{COM-as-Action} paradigm proposed by ComAct.

The three sub-corpora cover:

\begin{itemize}
    \item \textbf{com\_2d-20k}: 2D sketching and drafting COM operations
    (sketch constraints, dimensioning, drawing templates).
    \item \textbf{com\_3d-5k}: 3D part-modelling COM operations (sketch $\to$
    extrude / revolve / sweep / loft feature construction).
    \item \textbf{com\_assembly-5k}: assembly-level COM operations
    (\texttt{InsertMoveCopyBody2}, mate constraints, interference checks).
\end{itemize}

\paragraph{Source and construction.} All three sub-corpora are co-sourced with
data produced by ComAct, though we retain only the ``code IO'' portion for
SFT (dropping the agent-trajectory labels used by ComAct itself). Every raw
COM script has been \emph{sandbox-executed} in the \textbf{ComForge}
parallel-execution platform---each ran to completion inside a containerised
Windows VM instance of SolidWorks / Inventor / AutoCAD, produced concrete
CAD artefacts (\texttt{.stl} / \texttt{.step}), and only samples whose
execution succeeded \emph{and} whose artefacts passed structural checks were
kept. Data curated this way is far superior to the public-internet COM
scraps on both syntactic runability and engineering-semantic correctness. A more complete write-up of the COM-as-Action motivation
and the ComForge platform is available in the ComAct paper.


\subsubsection{Data statistics}
\label{sec:datastats}

\begin{figure}[t]
    \centering
    \includegraphics[width=0.9\columnwidth]{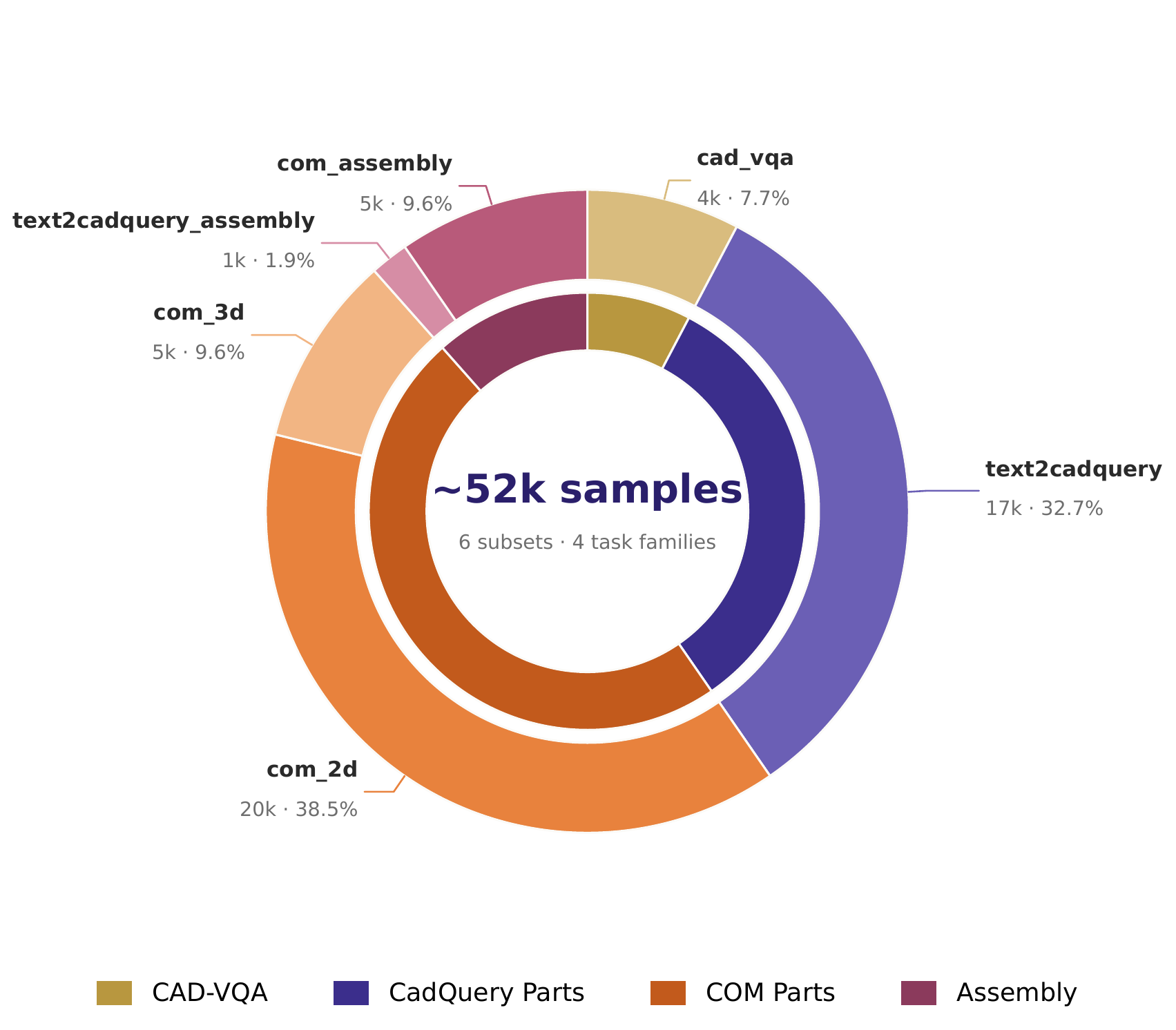}
    \caption{\textbf{Data distribution.} Outer ring: six sub-corpora sliced
    by sample count; consecutive slices belong to the same inner-ring family,
    giving visual alignment between the two levels. Inner ring: four task
    families (CAD-VQA, CadQuery Parts, COM Parts, Assembly). Two takeaways:
    (1)~the COM family (com\_2d + com\_3d) is the largest single-part
    contributor by sample count; (2)~the Assembly family is small in
    count~(6\,k) but each sample is long, so it contributes disproportionately
    to real training cost.}
    \label{fig:datadist}
\end{figure}

Table~\ref{tab:data-stats} aggregates the six sub-corpora into four task
families (CAD-VQA, CadQuery Parts, COM Parts, Assembly), matching the
inner-ring taxonomy of Figure~\ref{fig:datadist}, and
Figure~\ref{fig:datadist} shows both levels of aggregation---outer ring per
sub-corpus, inner ring per task family---sliced by sample count.

\begin{table}[t]
    \centering
    \small
    \caption{\textbf{Training data statistics.} Four task families over six
    sub-corpora, $\sim$52k samples in total. Rows follow the inner-ring
    ordering of Fig.~\ref{fig:datadist}; sub-corpus-level breakdowns appear in
    parentheses inside the ``Sub-corpora'' column.}
    \label{tab:data-stats}
    \begin{tabularx}{\columnwidth}{lXrlcl}
        \toprule
        Task family & Sub-corpora & Size & Primary modality & Avg.~tokens & Upstream source \\
        \midrule
        CAD-VQA        & cad\_vqa                                             &  4\,k & multi-view + text        & med   & Independent [TODO] \\
        CadQuery Parts & text2cadquery                                        & 17\,k & single-view + text       & med-L & IterCAD    \\
        COM Parts      & com\_2d (20\,k) + com\_3d (5\,k)                     & 25\,k & UI shot + text           & long  & ComAct     \\
        Assembly       & text2cadquery-assembly (1\,k) + com\_assembly (5\,k) &  6\,k & multi-view / UI + text   & long  & AssemCAD / ComAct \\
        \midrule
        \textbf{Total} & --- & $\sim$52\,k & --- & --- & --- \\
        \bottomrule
    \end{tabularx}
\end{table}

\subsubsection{Quality control and decontamination}
\label{sec:qc}

We apply two levels of decontamination to all six sub-corpora:
\textbf{(1)~UUID-level.} Every sub-corpus is precisely deduplicated against
every downstream benchmark (Zero-To-CAD, com\_QA\_mini, the CadQuery Assembly
evaluation set) by original-sample UUID---closing direct train/test leakage.
\textbf{(2)~Structural-level.} For CadQuery-related sub-corpora we further
hash on the AST-normalised script, closing trivial-variant leakage (variable
renaming, formatting differences) that would slip past UUID-level dedup.
The COM sub-corpora additionally apply a Chamfer-Distance nearest-neighbour
filter on the final CAD artefacts, keeping only samples with distinct
geometry to preserve coverage.

\subsection{Training recipe}
\label{sec:training-recipe}

\subsubsection{Base model choice}

We adopt \textbf{Qwen3.5-VL-27B}~\citep{qwen25vl} for three reasons:
(1)~the Qwen-VL family sits reliably in the top tier of public multimodal
baselines and its visual encoder handles grey-scale engineering line-work
adequately; (2)~a 27B parameter budget balances capability against
single-node $8{\times}$A100-80\,GB deployability, which downstream agents can
run uniformly; and (3)~Qwen3.5-VL's visual tokeniser gives us fine control
over image-token count via \texttt{IMAGE\_MAX\_TOKEN\_NUM}, providing stable
support for multi-view inputs.

\subsubsection{SFT rather than CPT}

The domain data already carries explicit QA / code pairs---the target side is
well-defined---so direct SFT is the more economical path. CPT is better
suited to scenarios where the target-language distribution itself must
migrate (e.g.\ bilingual CPT); this work concentrates on ``English code +
mixed-language descriptions $\to$ more structured CAD/COM code'', a typical
\emph{instruction-tuning-solvable} problem. From v1 onwards we therefore
run pure SFT.

\subsubsection{Key engineering settings}


\paragraph{Sequence parallel} (\texttt{sequence\_parallel\_size} $=8$).
    Long-sequence samples are sliced across 8 GPUs. CAD and COM scripts are
    much longer than typical instruction samples (\texttt{com\_assembly}
    routinely exceeds 5k tokens), and sequence parallel is the key to stable
    long-sample training. In early ablations, disabling it produced marked
    instability and memory pressure on long samples.

\paragraph{DeepSpeed ZeRO-3.} Optimiser states, gradients, and
    parameters are sharded within the data-parallel group. In conjunction
    with sequence parallel, a 27B multimodal SFT fits stably into a single
    $8{\times}$A100-80\,GB node.

\paragraph{Multi-task sampling ratio.} The six sub-corpora are mixed
    in a \emph{flat} proportion (no hand-tuned re-weighting). This is the
    simplest and most stable multi-task strategy.

\paragraph{Validation strategy.} Each sub-corpus maintains its own
    validation split, and we evaluate every sub-corpus independently during
    training. This prevents smaller sub-corpora (especially
    text2cadquery-assembly and CAD-VQA) from being diluted by the larger
    ones, and surfaces over- or under-fitting on the small tails early.

\section{Experiments}
\label{sec:experiments}

\subsection{Evaluation protocol}
\label{sec:protocol}

We partition evaluation into a \textbf{domain} suite and a \textbf{general}
suite, matching the multi-task structure of training.

\paragraph{Domain suite (4 benchmarks).}
\begin{itemize}
    \item \textbf{CAD~VQA} --- visual question answering over engineering
    drawings and 3D screenshots. Metric: QA accuracy.
    \item \textbf{CadQuery} --- parametric CAD code generation on
    single-part tasks (the Zero-To-CAD series). Metric:
    \texttt{passed@1e-3}, i.e.\ the fraction of samples whose generated code
    executes in the sandbox \emph{and} whose resulting geometry has a
    Chamfer Distance $\le 10^{-3}$ against the ground truth.
    \item \textbf{COM~CAD} --- Inventor / SolidWorks API code generation on
    the \texttt{com\_QA\_mini\_0311} evaluation set (60 samples). Metric:
    end-to-end pass rate of the final CAD artefact (sandbox execution +
    structural geometry match).
    \item \textbf{CadQuery~Assembly} --- assembly-level CAD code generation,
    sharing the evaluation set with AssemCAD. Same metric as CadQuery but
    with a higher bar: all parts of the assembly must match.
\end{itemize}

\paragraph{General suite (11 benchmarks).}
aime26, arc, chartqa, gsm8k, gsm8k\_v, math\_vision, mmlu, mmmu, ocr\_bench,
science\_qa, trivia\_qa. Coverage spans mathematics, multimodal reasoning,
charts, OCR, science QA, and commonsense. Each benchmark's primary metric
follows its published protocol (accuracy or pass rate).

\paragraph{Comparison models.} \qwenM (the untouched base model,
serving as the ``no-change baseline''); \gptM (a strong
closed-source frontier model representing ``the strongest publicly
consumable general model at the time of writing''); and
\ours (this work). \gptM was not evaluated
on the general suite, so those comparisons are two-way.

\paragraph{Evaluation infrastructure.} All four domain benchmarks share the
same \texttt{swift\_eval} sandbox-execution and Chamfer-Distance pipeline;
predictions, reviews, and aggregated reports live under
\texttt{swift\_eval/results/\{predictions,reviews,reports\}/\{model\}/\{bench\}...}. Aggregation is driven by \texttt{stat\_cd.py} and result management by
\texttt{btool.py}, which together keep results consistent across renames
and re-runs.

\subsection{Main results: domain benchmarks}
\label{sec:domain-results}

\begin{figure}[t]
    \centering
    \includegraphics[width=0.98\columnwidth]{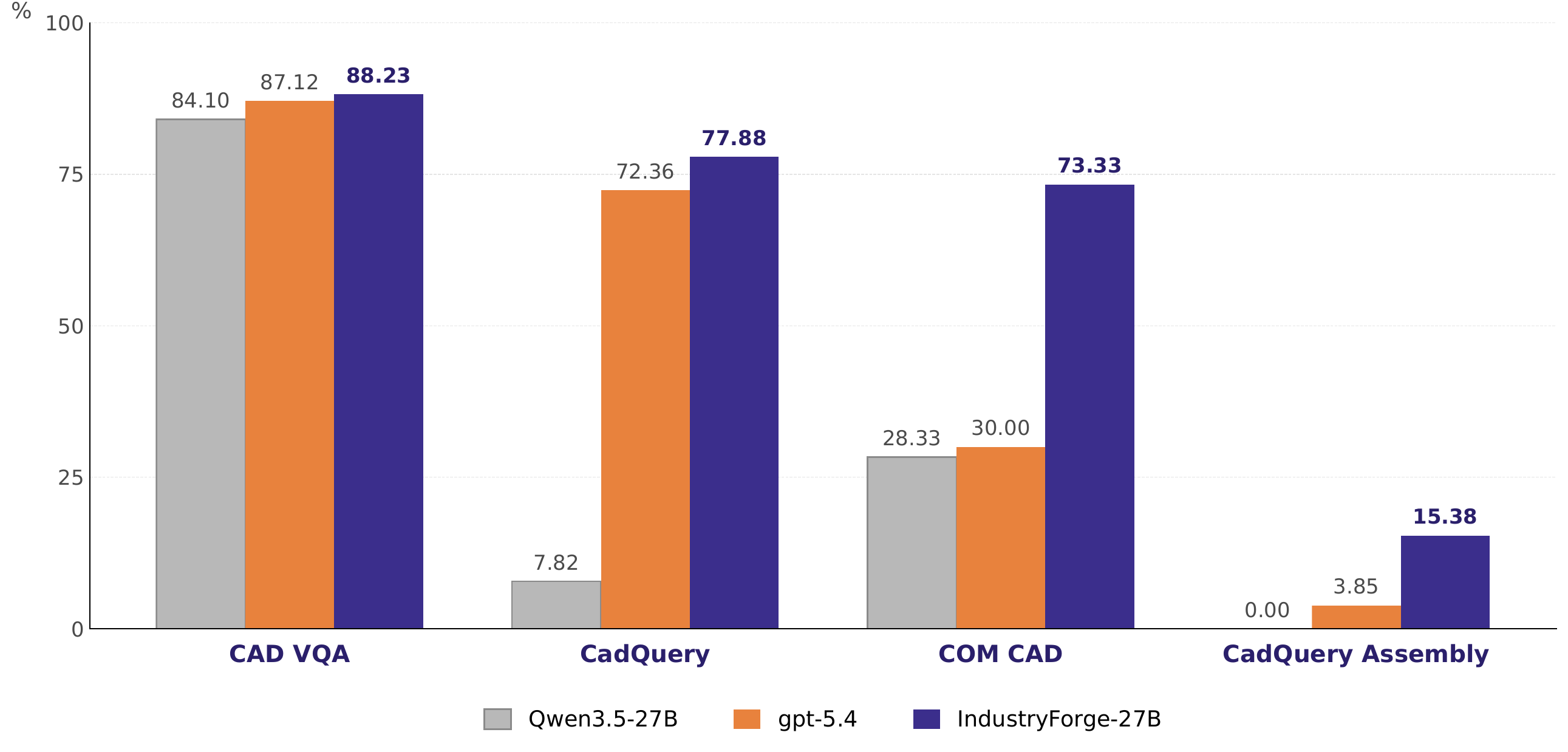}
    \caption{\textbf{Domain benchmark results.} Grouped bars per benchmark:
    \qwenM (grey), \gptM (orange),
    \ours (blue-purple). Value labels above each bar.
    Mean $= 63.71\%$, $+33.65$~pp over the base model, and \ours
    wins on 4/4 benchmarks against \gptM.}
    \label{fig:domain}
\end{figure}

Figure~\ref{fig:domain} summarize the domain results.

\paragraph{Key observations.}
\begin{itemize}
    \item \textbf{An order-of-magnitude jump on CadQuery.}
    $7.82\% \to 77.88\%$ ($+70.06$~pp), \emph{overtaking} \gptM
    (72.36\%)---strong evidence that domain SFT has enormous leverage on
    parametric CAD code generation.
    \item \textbf{Crushing the closed-source model on COM CAD.}
    $28.33\% \to 73.33\%$ ($+45.00$~pp), and $+43.33$~pp ahead of \gptM.
    COM code is exactly where public data are scarcest, and it is also where
    the closed-source model shows the biggest gap---making this the largest
    source of our differentiation.
    \item \textbf{First usable non-zero assembly-level pass rate.} Both the
    base model and \gptM sit in the near-total-failure regime
    (0.00\% / 3.85\%). \ours reaches 15.38\%---to our knowledge,
    the first releasable industrial-CAD foundation with a non-zero usable
    pass rate on assembly-level generation.
\end{itemize}

\subsection{Main results: general benchmarks}
\label{sec:general-results}

Table~\ref{tab:general} presents the 11-benchmark general suite, grouped by
capability area (Mathematics / Knowledge Reasoning / Document Understanding)
and, within each group, alphabetically. $\Delta$ against the base model is
discussed in the prose analysis rather than tabulated.

\begin{table}[t]
    \centering
    \small
    \caption{\textbf{Main results on general benchmarks.} Best cells in
    bold. Groups: Mathematics / Knowledge Reasoning / Document Understanding.}
    \label{tab:general}
    \begin{tabular}{lccc}
        \toprule
        Benchmark          & \qwenM     & \gptM          & \ours \\
        \midrule
        \multicolumn{4}{l}{\textit{Mathematics}} \\
        aime26             & \textbf{83.33\%} & 66.67\%          & \textbf{83.33\%} \\
        gsm8k              & 96.29\%          & \textbf{97.60\%} & 97.35\%          \\
        gsm8k\_v           & \textbf{57.24\%} & 44.60\%          & 52.99\%          \\
        math\_vision       & 47.33\%          & \textbf{66.20\%} & 62.30\%          \\
        \midrule
        \multicolumn{4}{l}{\textit{Knowledge Reasoning}} \\
        arc                & 95.15\%          & 95.20\%          & \textbf{95.30\%} \\
        mmlu               & 90.14\%          & \textbf{92.77\%} & 90.95\%          \\
        mmmu               & 78.33\%          & 77.00\%          & \textbf{78.56\%} \\
        science\_qa        & \textbf{95.99\%} & 94.80\%          & 94.84\%          \\
        trivia\_qa         & 75.88\%          & 78.00\%          & \textbf{79.12\%} \\
        \midrule
        \multicolumn{4}{l}{\textit{Document Understanding}} \\
        chartqa            & 80.56\%          & 76.20\%          & \textbf{84.40\%} \\
        ocr\_bench         & \textbf{90.30\%} & 78.90\%          & 88.60\%          \\
        \midrule
        \textbf{Mean (11)} & 81.86\%          & 78.90\%          & \textbf{83.42\%} \\
        \bottomrule
    \end{tabular}
\end{table}

Across the 11 general benchmarks, \ours improves the base model
on 7 with a mean $+1.56$~pp---no catastrophic forgetting. Notably, the
capabilities most correlated with what our CAD/COM corpus emphasises---%
\emph{multimodal geometric reasoning}, \emph{structured document understanding},
and \emph{long-context / multi-step consistency}---all see positive spillover:

\begin{itemize}
    \item \textbf{\texttt{math\_vision} $+14.97$~pp}, the single largest
    gain. \texttt{math\_vision} draws heavily on ``read three views / spatial
    diagrams $\to$ reason about length / area / angle / volume''---a mental
    process highly homologous to CAD-VQA's ``three views $\to$ structured
    dimension QA'' and to text2cadquery's ``single- or multi-view $\to$
    parametric geometry script''. The internal ``multi-view $\to$ geometry''
    representation acquired during SFT transfers directly to
    \texttt{math\_vision}-style geometric-visual reasoning.

    \item \textbf{\texttt{chartqa} $+3.84$~pp}. CAD engineering drawings
    embed dense annotations and tabular dimension lists whose visual
    structure is homologous to chart QA. The model's SFT-acquired ability to
    ``read numbers from compact figure/table structure'' spills over onto
    chart-QA.

    \item \textbf{\texttt{trivia\_qa} $+3.24$~pp and \texttt{gsm8k}
    $+1.06$~pp}. CAD and COM code corpora are archetypal \emph{long-context,
    multiple variable bindings, structured output} samples; they strengthen
    the model's grip on long-context consistency and multi-step symbolic
    stability, which shows up as light but consistent gains on tasks that
    demand cross-sentence factual and numerical consistency.

    \item \textbf{\texttt{mmlu} $+0.81$~pp, \texttt{mmmu} $+0.23$~pp,
    \texttt{arc} $+0.15$~pp}. Three knowledge-reasoning benchmarks all
    register small positive gains. The absolute magnitudes are modest but
    they demonstrate that injecting CAD / COM data does \emph{not} squeeze
    out the model's existing academic and commonsense knowledge---a direct
    dividend of a conservative multi-task-SFT recipe (LoRA, frozen visual
    tower).
\end{itemize}

Taken together, the seven positive shifts cluster along three axes---%
\emph{geometric visual reasoning}, \emph{structured figure-text
understanding}, and \emph{long-context / multi-step reasoning}. They are
not a specialisation-for-generalisation trade-off but a secondary spillover
naturally carried by the design of a CAD / COM SFT corpus, and lend
side-evidence to our decision to pre-install ``vision + structured code''
capabilities into the foundation-model layer.






\subsection{Case studies}
\label{sec:case-studies}

\begin{figure}[t]
    \centering
    \includegraphics[width=0.98\columnwidth]{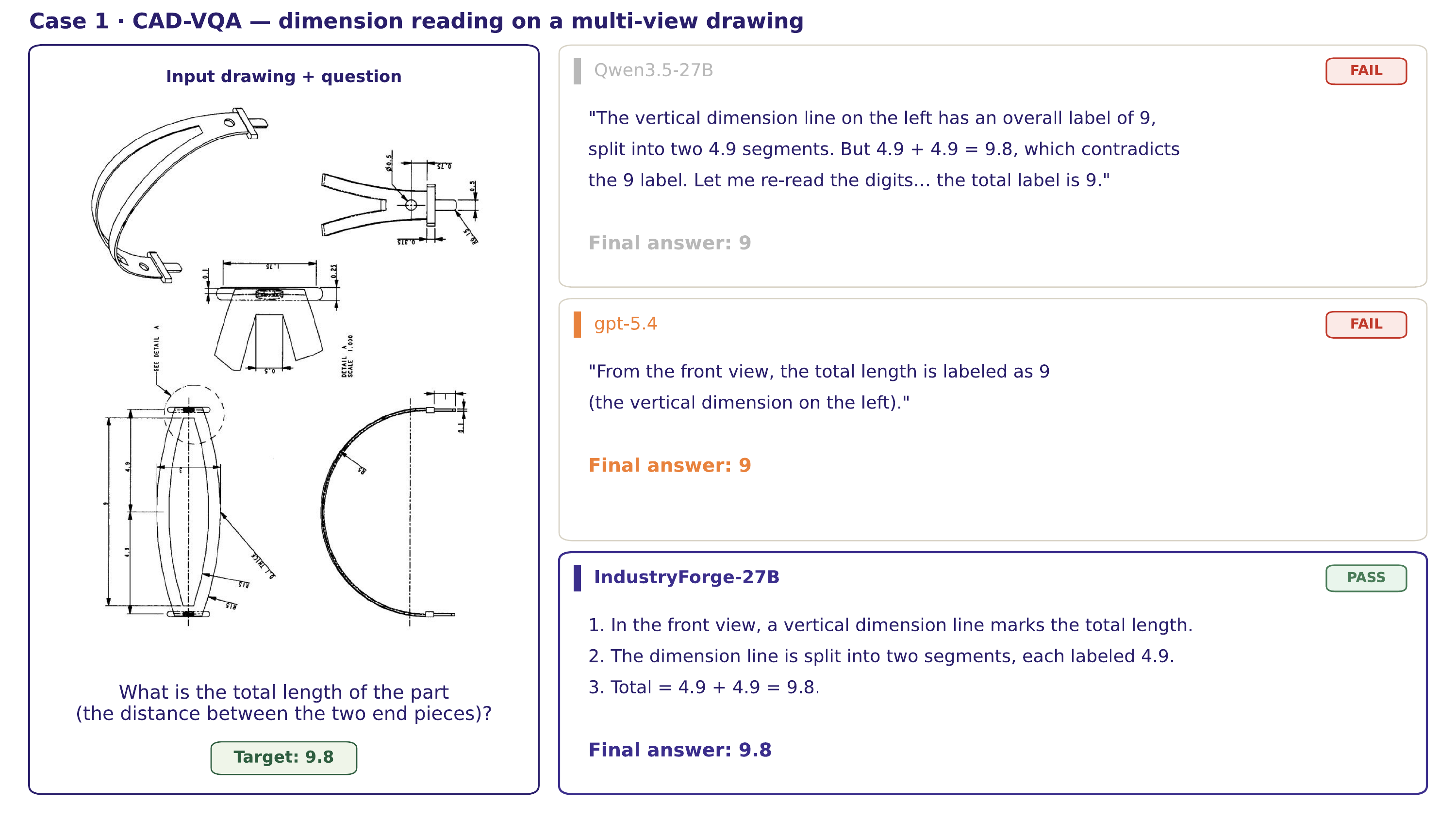}
    \caption{\textbf{Case~1 --- CAD VQA (part identification / dimension
    reading).} Prose-style QA layout: an input image and question on the
    left; three stacked answer rows on the right showing each model's
    chain-of-thought excerpt and final answer with PASS / FAIL badges.
    \ours correctly reads the multi-view geometry and matches
    the target answer; the base model and \gptM mis-count features.}
    \label{fig:case1}
\end{figure}

\begin{figure}[t]
    \centering
    \includegraphics[width=0.98\columnwidth]{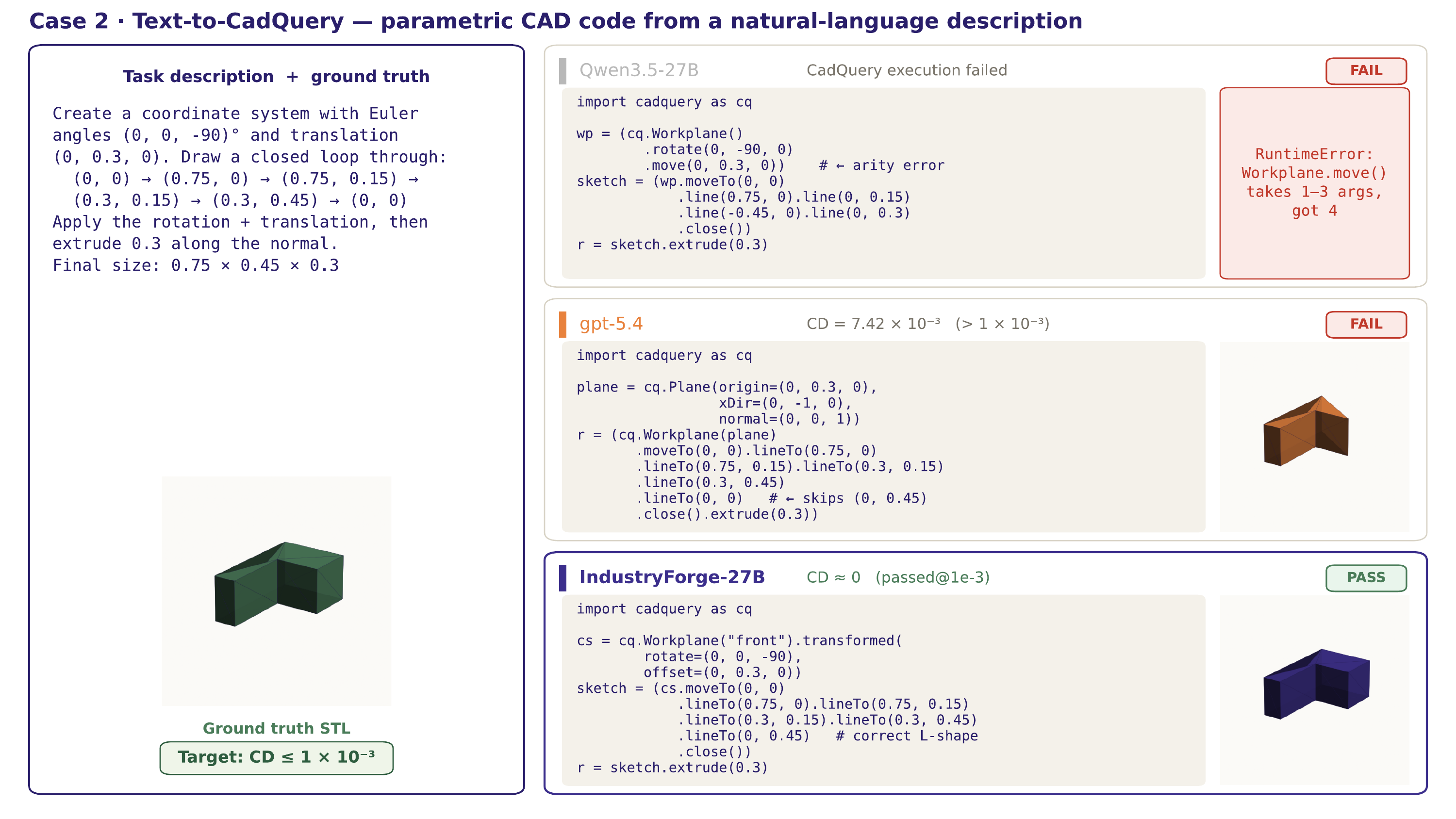}
    \caption{\textbf{Case~2 --- Text-to-CadQuery (single part).} Left card:
    task description, ground-truth STL preview, and CD~$\le 10^{-3}$ target.
    Right: three stacked rows with per-model code excerpts (the discriminating
    lines highlighted) and the corresponding STL renders.
    \ours produces a CadQuery script that passes at $10^{-3}$;
    the base model errors out on the sketch API and \gptM mis-dimensions
    the target feature.}
    \label{fig:case2}
\end{figure}

\begin{figure}[t]
    \centering
    \includegraphics[width=0.98\columnwidth]{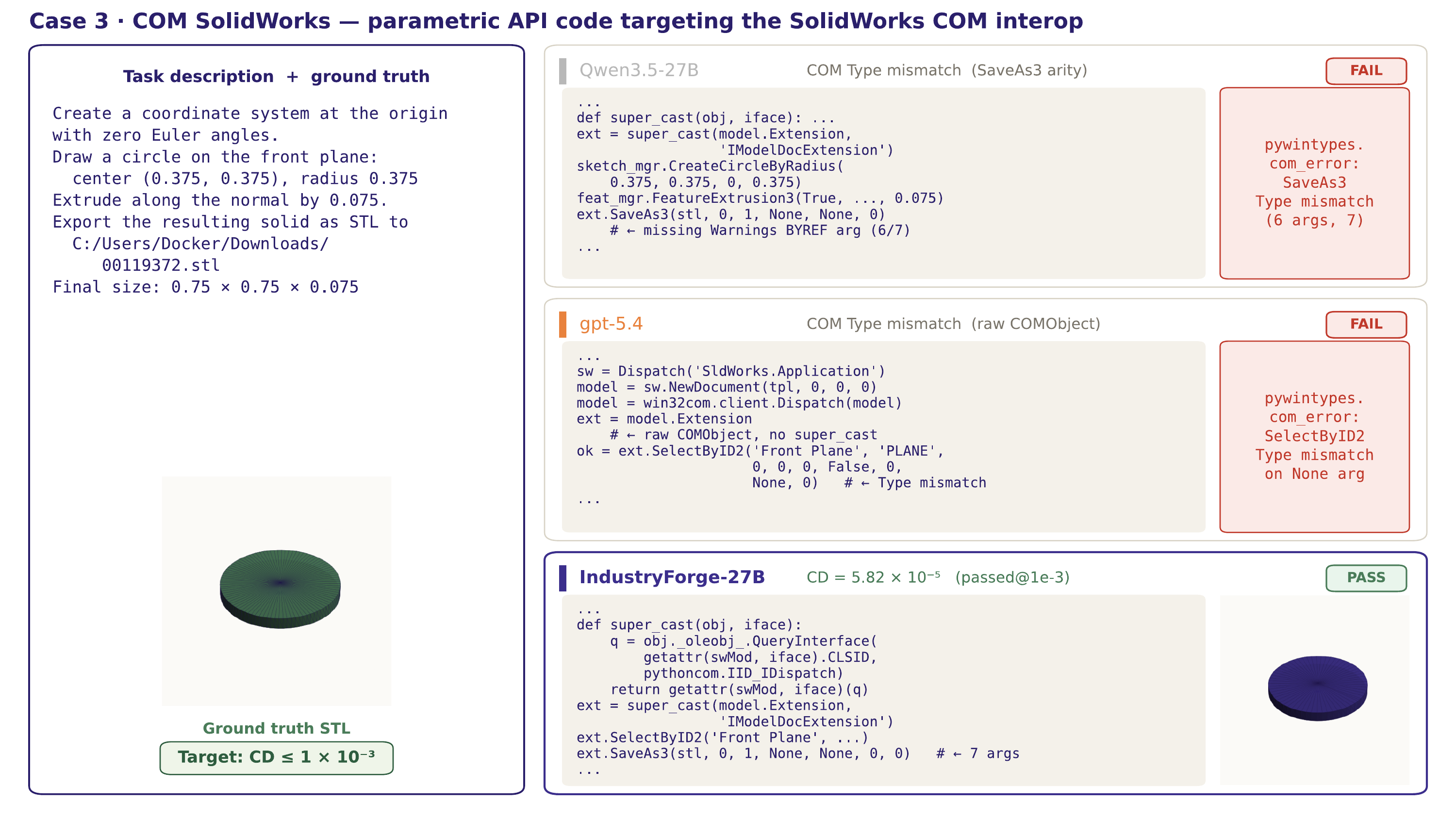}
    \caption{\textbf{Case~3 --- COM (SolidWorks).} A single-cylinder task
    executed through the SolidWorks COM API. Each row shows the model's
    key COM calls (sketch $\to$ extrude $\to$ move body) and the resulting
    STL. \ours uses the correct helper structure and passes;
    \gptM mis-selects the extrude direction; the base model fails at COM
    plane construction.}
    \label{fig:case3}
\end{figure}

\begin{figure}[t]
    \centering
    \includegraphics[width=0.98\columnwidth]{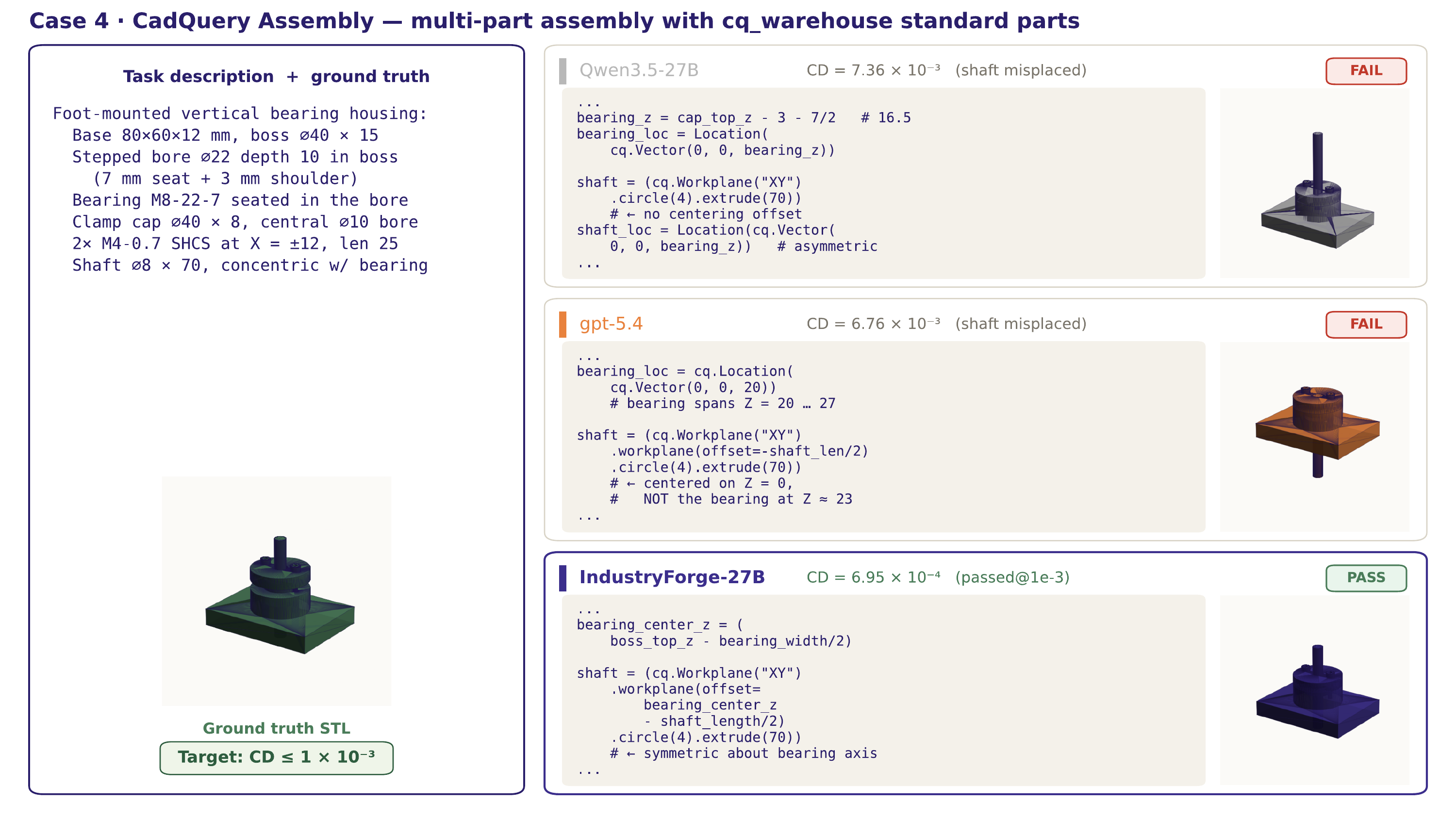}
    \caption{\textbf{Case~4 --- CadQuery Assembly (multi-part with
    \texttt{cq\_warehouse}).} A foot-mounted vertical bearing housing with
    clamp cap and shaft. The discriminating detail is a single line of
    coordinate reasoning that centres the shaft on the bearing axis rather
    than on the global origin or on the bearing's local upward extrude.
    Only \ours's placement (\texttt{offset =
    bearing\_center\_z $-$ shaft\_length/2}) passes at $10^{-3}$.}
    \label{fig:case4}
\end{figure}

\begin{figure}[t]
    \centering
    \includegraphics[width=0.98\columnwidth]{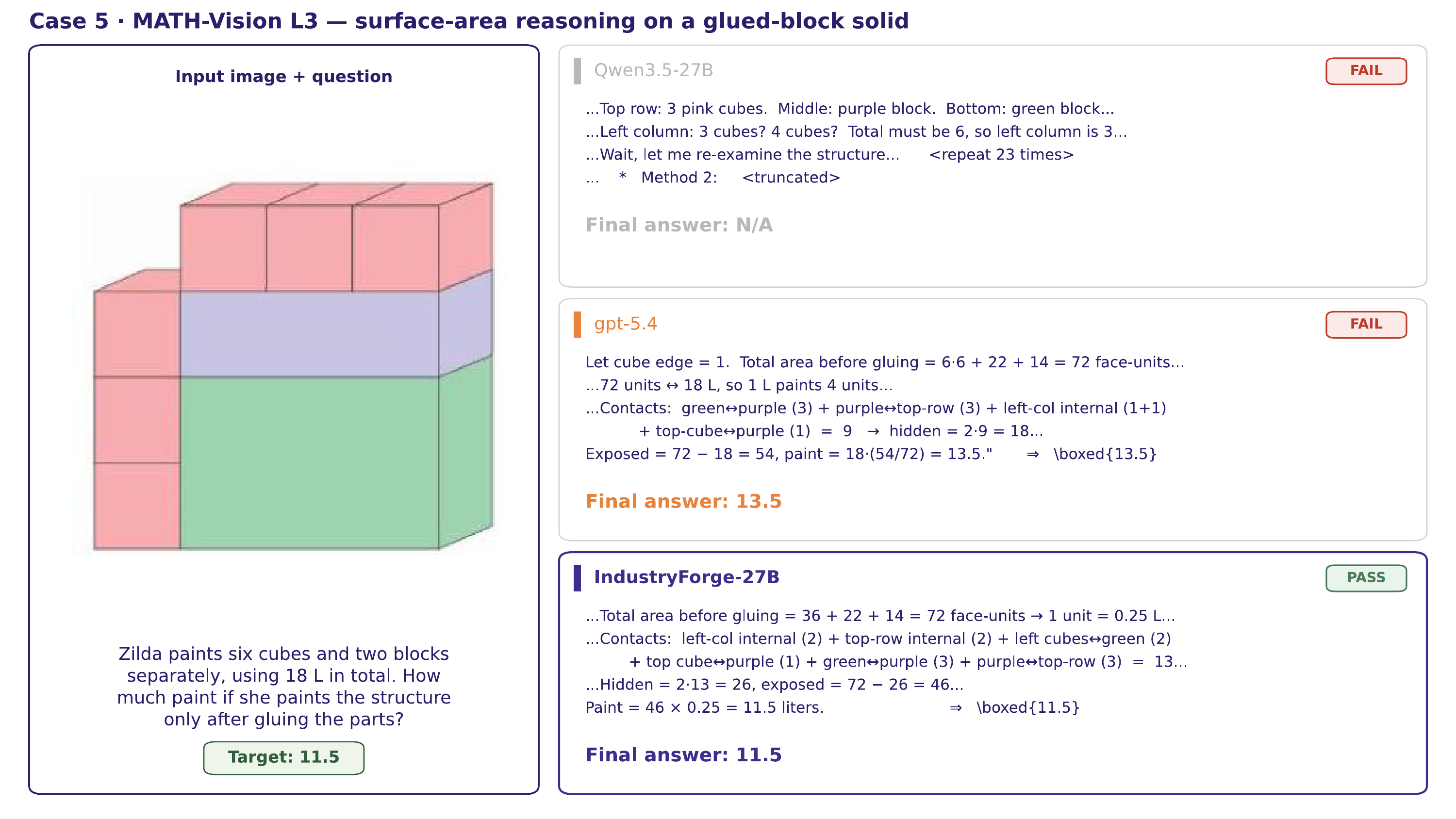}
    \caption{\textbf{Case~5 --- Cross-benefit on math\_vision L3.} A
    painted-solid surface-area puzzle. The base model runs into a
    ``rathole'' (23 rounds of ``\emph{Wait, let me re-examine}\ldots'' over
    40k characters, the correct value 11.5 appearing at 92\% of the response
    but never committed to \texttt{\textbackslash boxed\{\}}); \gptM
    under-counts contact interfaces and lands on 13.5; \ours
    enumerates all six contact groups, computes hidden $=26$, exposed $=46$,
    and commits to 11.5. This is direct evidence that CAD SFT lifts
    ``multi-view $\to$ spatial geometry'' reasoning on general benchmarks.}
    \label{fig:case5}
\end{figure}

We close the experiments with five qualitative case studies that make the
domain / general story concrete (Figure~\ref{fig:case1}-Figure~\ref{fig:case5}). Each case contrasts \qwenM, \gptM,
and \ours on the same input.

\section{Conclusion}
\label{sec:conclusion}

Starting from Qwen3.5-VL-27B, this work
performs multi-task SFT on six industrial-domain multimodal sub-corpora
($\sim$52k samples in total) to produce \ours---a foundation
model for industrial agents that covers four core skills: CAD visual
understanding, parametric CAD code generation, assembly-level CAD code
generation, and COM industrial-software code generation. On the four
CAD-domain benchmarks the model averages $+33.65$~pp over its base and wins
4/4 against \gptM; on 11 general-capability benchmarks it averages
$+1.56$~pp with no catastrophic forgetting.

\ours is \emph{not} an end-to-end CAD
generation system, but the \emph{common substrate} for industrial-agent
projects. It settles the prerequisite question ``can our foundation model
read CAD drawings and write CAD code at all?'' once at the foundation-model
layer, so that IterCAD, AssemCAD, ComAct, and SimLoop all \emph{will} start
from the same launchpad. Because none of these downstream projects has yet
formally integrated the substrate, we use the future tense throughout this
report when describing that role.

\paragraph{Limitations.}
\begin{itemize}
    \item Single-part samples still dominate the training distribution;
    assembly-level pass rate, though lifted from 0 to a usable non-zero
    (15.38\%), has substantial room to improve.
    \item The current scope only covers CAD code generation and visual
    understanding; simulation and CAE are not modelled at the foundation
    layer, and depend on SimLoop's CAD-CAE loop to close that gap.
    \item This report covers only SFT---closed-loop RL is deferred to the
    IterCAD line of work on top of the substrate.
    \item The COM sub-corpora focus on the ``Windows CAD trio'' (SolidWorks,
    Inventor, AutoCAD). Other COM domains (Office, Adobe) are not yet
    included.
\end{itemize}



\clearpage
{
\bibliographystyle{unsrt}  
\bibliography{preprint}
}



\end{document}